\documentclass[letterpaper]{article} 
\usepackage[]{aaai25}  
\usepackage{times}  
\usepackage{helvet}  
\usepackage{courier}  
\usepackage[hyphens]{url}  
\usepackage{graphicx} 
\usepackage{natbib}  
\usepackage{caption} 
\usepackage{booktabs}
\usepackage{algorithm}
\usepackage{algorithmic}
\usepackage{multirow}
\usepackage{microtype}
\usepackage{amsmath,amsfonts,amssymb}
\usepackage{upgreek}
\usepackage{newfloat}
\usepackage{listings}
\DeclareCaptionStyle{ruled}{labelfont=normalfont,labelsep=colon,strut=off} 
\floatstyle{ruled}
\newfloat{listing}{tb}{lst}{}
\floatname{listing}{Listing}
\title{DiffCLIP: Few-shot Language-driven Multimodal Classifier}
\author{
    Jiaqing Zhang$^{1}$, 
    Mingxiang Cao$^{1}$,
    Xue Yang$^{2}$,
    Kai Jiang$^{1}$,
    Yunsong Li$^{1}$,
}
\affiliations{

    $^1$The State Key Laboratory of Integrated Services Networks, Xidian University \\ $^2$Shanghai AI Laboratory

}

\usepackage{bibentry}

\begin{document}

\maketitle

\begin{abstract}
Visual language models like Contrastive Language-Image Pretraining (CLIP) have shown impressive performance in analyzing natural images with language information. However, these models often encounter challenges when applied to specialized domains such as remote sensing due to the limited availability of image-text pairs for training. To tackle this issue, we introduce DiffCLIP, a novel framework that extends CLIP to effectively convey comprehensive language-driven semantic information for accurate classification of high-dimensional multimodal remote sensing images. DiffCLIP is a few-shot learning method that leverages unlabeled images for pretraining. It employs unsupervised mask diffusion learning to capture the distribution of diverse modalities without requiring labels. The modality-shared image encoder maps multimodal data into a unified subspace, extracting shared features with consistent parameters across modalities. A well-trained image encoder further enhances learning by aligning visual representations with class-label text information from CLIP. By integrating these approaches, DiffCLIP significantly boosts CLIP performance using a minimal number of image-text pairs. We evaluate DiffCLIP on widely used high-dimensional multimodal datasets, demonstrating its effectiveness in addressing few-shot annotated classification tasks. DiffCLIP achieves an overall accuracy improvement of 10.65\% across three remote sensing datasets compared with CLIP, while utilizing only 2-shot image-text pairs.
\end{abstract}

%

\section{Introduction}
Remote sensing images captured over the same geographic region by different sensors often provide complementary ground features \cite{zhang2024multimodal}. Joint classification of multimodal remote sensing data leverages the integration of these complementary sources, enhancing classification accuracy. This approach has been extensively applied in various domains, including urban planning \cite{dong2023abundance}, natural resource management \cite{wu2021convolutional}, and environmental monitoring \cite{dong2022multibranch}, among others. A typical high-dimensional remote sensing image, hyperspectral imaging (HSI), provides rich high-dimensional spectral information, which can be used for material identification based on reflectance values, thus providing new multi-dimensional modal information for remote sensing image classification \cite{roy2023spectral}. Effectively utilizing high-dimensional spectral information to integrate and learn features from different modalities for better understanding and representation of cross-modal features becomes the key to high-dimensional multimodal joint representation learning. Although each modality has unique features, they often share common information in the semantic space. For example, Hazarika \textit{et al.} \cite{hazarika2020misa} introduced a shared subspace to discover potential commonalities between different modalities, aiming to reduce the impact of the modality gap. Dutt \textit{et al.} \cite{dutt2022shared} developed a universal shared manifold model, which can learn shared feature representations from hyperspectral and light detection and ranging (LiDAR) images. Subsequently, the transformer \cite{roy2023multimodal} and diffusion model \cite{zhou2023hyperspectral} take the classification model to a larger scale. \textbf{However, these models only focus on seeking consistency within the latent semantic space in the visual image dimension, lacking joint exploration based on the visual-language view.}

\begin{figure}[tpb]
    \centering
    \includegraphics[width=\linewidth]{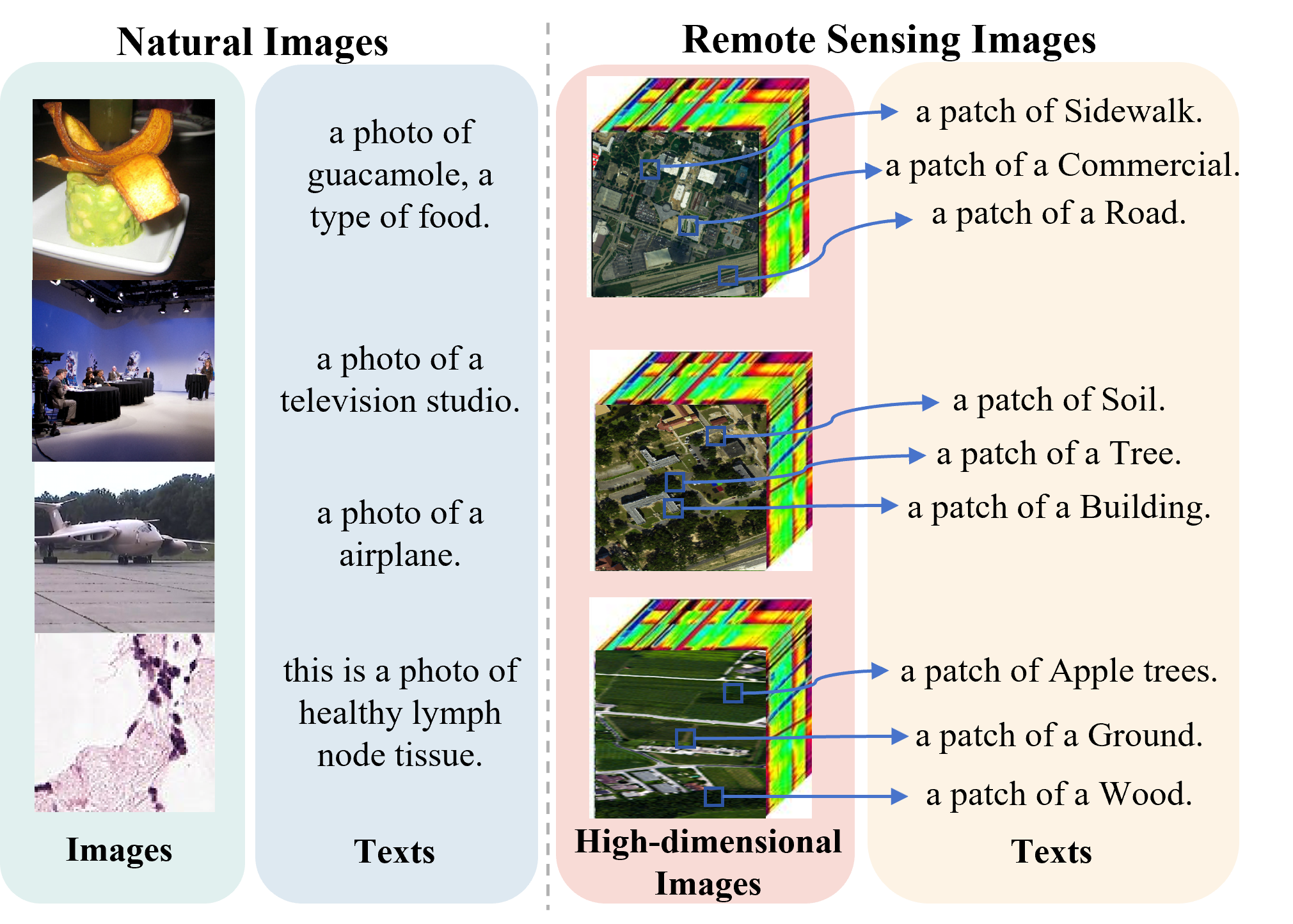}
    \caption{(a) The CLIP model is trained using four randomly sampled image-text pairs from natural image datasets, which are rich in labeled examples.(b) Remote sensing images are annotated in patch format, but there is a severe scarcity of annotated patch samples due to the specialized nature of remote sensing annotations, requiring professional expertise and efficient time management. This significant disparity between datasets makes it challenging to directly apply the CLIP model to remote sensing applications.}
    \vspace{-0.1in}
    \label{figure1}
    \vspace{-0.1in}
\end{figure}

Recently, the contrastive language image pre-training framework (CLIP) \cite {radford2021learning} has achieved remarkable success, providing a foundation for many subsequent tasks such as semantic segmentation \cite {rao2022denseclip}, object detection \cite{shi2023edadet}, and 3D point cloud understanding \cite{zhang2022pointclip}. The latest progress in CLIP models mainly focuses on scaling the model size and data size \cite{cherti2023reproducible}, combining self-supervision \cite{mu2022slip, li2021supervision, yu2022coca}, improving pre-training efficiency \cite {chen2023protoclip}, and few-shot adaptation \cite{zhou2022learning}. However, these models often struggle when applied to specific fields such as remote sensing images, especially high-dimensional data as shown in Figure \ref{figure1}. This is because these models are trained on natural images, which may not fully capture the diversity and complexity of specific fields. To address this issue, most research has focused on constructing large-scale pre-training datasets for each domain, with additional fine-tuning stages to adapt to downstream tasks in medical \cite{zhang2022contrastive, PubMedCLIP,wang-etal-2022-medclip}, e-commerce \cite{dong2022m5product,liu2023mep} and remote sensing \cite{zhang2023rs5m,liu2023remoteclip} field. However, the requirement for professional availability for large-scale high-dimensional image datasets constrains supervised learning of high-dimensional images. Therefore, a natural question arises: \textbf{can we avoid the cost of collecting and labeling data and introduce CLIP into high-dimensional multimodal image classification with fewer labeled samples?} Unfortunately, the use of unsupervised learning knowledge to handle CLIP few-shot training has not been fully explored.

To overcome this issue, we propose DiffCLIP, a novel method that provides a few-shot training paradigm for high-dimensional multimodal remote sensing image classification. We begin by employing mask diffusion unsupervised learning to capture the distributions of various visual image modalities without relying on labels. To enhance the model's ability to extract semantic information, we introduce mask operations and diffusion processes. This operation facilitates sparse representation of input data, reducing interference from redundant information in multimodal remote sensing data, and accelerating training speed. By utilizing a visual image encoder with shared parameters, we map multimodal data into a shared subspace and conduct mask restoration through two modality-specific decoders. This reconstructs input visual image modalities to capture specific attributes of different modalities. The entire encoding and decoding process is integrated into a denoising diffusion model with strong implicit learning capabilities, helping to deal with the difficulty of reconstruction caused by the huge modal gap.

Language-driven few-shot classification is a text feature-driven supervised learning process, where a well-trained visual image encoder enables CLIP to effectively utilize few-shot class label text information for supervision. The semantic information of different modalities is aligned with the class label text information obtained from the text encoder, promoting consistent learning of visual representations across different modalities. Compared to discrete label values, it provides a more comprehensive semantic information representation. Leveraging language-driven methods helps the model capture rich inherent semantic details in complex data distributions, thereby enhancing classification performance.

Extensive experiments on several downstream tasks demonstrate the effectiveness of the proposed DiffCLIP. Specifically, on the Houston dataset, using unlabeled patch samples as the unsupervised mask diffusion pre-training dataset and utilizing the transformer as the image text encoder, DiffCLIP achieved an overall accuracy of 52.15\% on 2-shot classification tasks, which is 16.32\% higher than the baseline CLIP directly pre-trained with ViT-B-14. Our contributions are summarised as follows:
\begin{itemize}
    \item We provide a few-shot training paradigm for a high-dimensional multimodal image classification framework, by exploring the potential of CLIP in specific few-shot learning domains.
    \item We design an unsupervised mask diffusion process to extract shared features across multiple modalities, providing a robust visual image encoder for CLIP to be introduced into the specialized domain of few-shot learning.
    \item We propose a language-driven framework that introduces class-label text information to enhance the extraction of semantic information of the multimodal visual image encoder, thereby encouraging consistent learning of visual representations across different modalities.
\end{itemize}

\begin{figure*}[tp]
    \centering
    \includegraphics[width=\linewidth]{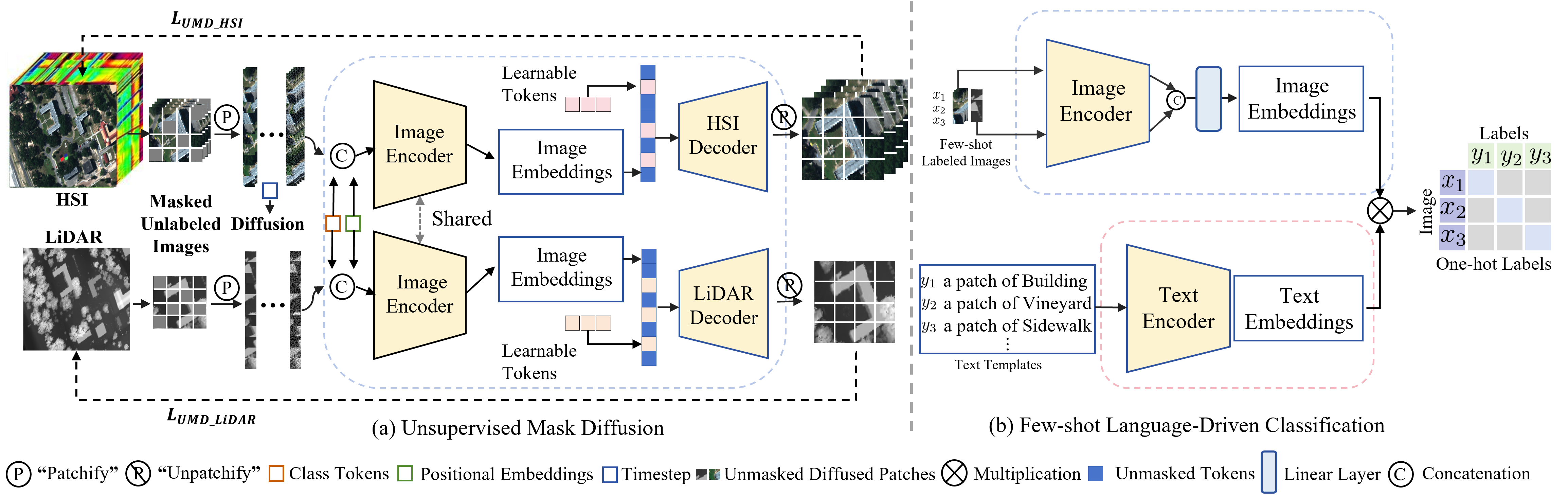}
    \caption{The DiffCLIP framework consists of two main stages: a) Unsupervised Mask Diffusion: A modality-shared image encoder captures consistent features across two modalities, while two modality-specific decoders integrate semantic prompts and unique features. b) Few-shot Language-Driven Classification: DiffCLIP fine-tunes the modality-shared encoder and employs language methods to convey comprehensive semantic information. This approach helps capture rich semantic information inherent in complex data distributions.}
    \vspace{-0.1in}
    \label{fig:framework}
    \vspace{-0.1in}
\end{figure*}

\section{Related Work}
\subsection{Vision-language pre-training (VLP)}  
Pre-trained visual language models (VLMs) have achieved remarkable success, serving as the basis for various downstream tasks \cite{bommasani2021opportunities}. Existing methods include reconstructing masked inputs \cite{chen2020uniter,kim2021vilt}, learning joint visual-language embeddings \cite{yu2022coca}, generating language descriptions from image inputs \cite{yu2022coca}, and linking pre-trained unimodal image and language models \cite{Desai_Johnson_2021}. CLIP \cite{radford2021learning}, a leading model for joint visual-language embeddings, has demonstrated effectiveness across various applications \cite{radford2021learning, Luo_Ji_Zhong_Chen_Lei_Duan_Li_Empirical_Lei,Gu_Lin_Kuo_Cui_2021}. 
Despite training on large numbers of natural image samples, VLMs exhibit difficulty capturing diversity and complexity in professional fields like remote sensing \cite{ kim2023bias}, hindering direct deployment in such domains. Data in these fields is often scarce, with limited accessibility and requiring expert annotations, making obtaining image-text pairs for VLM training challenging. 

\subsection{Few-shot Learning}
Few-shot learning \cite{li2020boosting} aims to enable models to learn and adapt effectively to new tasks with limited labeled samples in unseen classes. Traditional few-shot learning falls into two categories: meta-learning-based \cite{santoro2016meta,finn2017model} and transfer learning-based \cite{tian2020rethinking,li2022cross} methods. In remote sensing, few-shot learning has been successfully applied in the remote sensing fields \cite{liu2018deep,zhang2022graph,dai2024learning,wang2024dual}. However, current few-shot methods primarily focus on single-modality tasks and lack discussions on multimodal applications. CLIP demonstrates impressive few-shot classification capabilities by pre-training on a large set of image-text pairs. Recent advancements \cite{zhu2023not,song2022clip,zhang2022tip} have utilized CLIP’s multimodal capabilities to tackle few-shot challenges. For instance, CoOp \cite{zhou2022learning} and CoCoOp \cite{zhou2022conditional} introduced methods for automated prompt engineering to enhance prompt adjustments, which typically require domain expertise and are time-consuming. 
Despite the effectiveness of CLIP, their adaptation to specific domains, especially in data-limited scenarios like few-shot remote sensing tasks, remains understudied. This paper proposes a language-driven few-shot classification model based on diffusion unsupervised learning to address limited training samples and high transfer difficulties when applying CLIP to specialized fields.



\section{Method}
The proposed DiffCLIP is a few-shot learning method aimed at learning shared features across different modalities through unsupervised training of an image encoder using mask diffusion, followed by matching the representations of image features guided by text semantic supervision. This method addresses high-dimensional multimodal few-shot remote sensing image classification problems. As depicted in Figure \ref{fig:framework}, our method comprises two main processes: unsupervised learning and few-shot learning. The unsupervised learning stage involves two parts: forward mask diffusion and reverse denoising restoration. To lower the training cost of the diffusion model, we employ a strategy where only a subset of the training dataset is sampled, and forward mask diffusion is applied. In the reverse denoising restoration, a shared image encoder is employed to learn multimodal shared features. Additionally, two modality-specific decoders are designed to capture specific attributes of different modalities. The few-shot learning stage is a language-driven supervised learning process, which offers a more comprehensive representation of semantic information compared to discrete label values. This richer semantic context assists the model in capturing the nuanced details present in complex data distributions, thereby enhancing classification performance. Additionally, for the sake of brevity, in most cases, we only present the formulas for single-modal inputs, as they are generally applicable to both modalities.
\subsection{Unsupervised Mask Diffusion}
\subsubsection{Mask Diffusion Process}
Given a clean sample $x_0\sim \mathcal{Q}(x_0)$ in the forward mask diffusion, DiffCLIP employs an asymmetric masking strategy \cite{he2022masked} for each modality, to encourage the model to effectively capture shared features across different modalities.
The $x_{0}$ is firstly divided into the non-overlapping masked region $x_{0}^{m}$ according to a fixed masking ratio and treat the rest as visible patches $x_{0}^{v}$. Only the visible area $x_{0}^{v}$ is gradually diffused, corrupted by recursively adding a small amount of Gaussian noise $T$ times with variance $\beta_{t}\in(0,1)$ to produce $x_{1}^{v}$, $x_{2}^{v}$, \ldots, $x_{T}^{v}$ following the Markov process below:
\begin{equation} \label{3.1}
    \mathcal{Q}(x_t^v|x_{t-1}^v)=\mathcal{N}(x_t^v;\sqrt{1-\beta_t}x_{t-1}^v,\beta_tI),
\end{equation} 
\begin{equation} \label{3.2}
    \mathcal{Q}(x_1^v,\ldots,x_T^v|x_0^v)=\prod_{t=1}^T\mathcal{Q}(x_t^v|x_{t-1}^v),
\end{equation} 
where $t$ $\in$ [1, 2, \ldots, $T$] denotes the timestep, $\beta_{1:T}$ is pre-defined and undergoes a gradual linear decay, held as hyper-parameters. The mask diffusion operation creates a task that cannot be easily solved by extrapolating visible neighboring patches (due to adding noise to visible patches), while also minimizing redundant information and resulting in highly sparse inputs. This helps reduce the computational cost of the diffusion model.
\subsubsection{Denoising Restoration Process}
During the reverse process, DiffCLIP predicts input data $x_{0}$ based on the current sampling time $t$. This modification is based on the Bayesian theory as follows:
\begin{equation} \label{3.3}
    \mathcal{Q}(x_{t-1}|x_{t},x_{0})=\mathcal{N}(x_{t-1};\tilde{\mu}_{t}(x_{t},x_{0}),\tilde{\beta}_{t}I),
\end{equation} 
when the variance of noise $\beta_t$ in each step $t$ is small enough, $\mathcal{Q}(x_{t-1}|x_{t},x_{0})$ can also be considered Gaussian distributed, approximated by a deep network as follows:
\begin{equation} \label{3.4}
    \mathcal{P}_\theta(x_{t-1}^v|x_t^v)=\mathcal{N}(x_{t-1}^v;\mu_\theta(x_t^v,t),\Sigma_\theta(x_t^v,t)),
\end{equation} 
where $\mu_\theta(x_t^v,t)$ is the mean of $x_t^v$, $\Sigma_\theta(x_t^v,t)$ is the variance of $x_t^v$, and the joint distribution $\mathcal{P}_\theta(x_{0:T}^v)$ is defined as a Markov chain with learned Gaussian transitions starting at $\mathcal{P}(x_{T}^v)=\mathcal{N}(x_{T}^v;0,I)$:
\begin{equation} \label{3.5}
    \mathcal{P}_\theta(x_{0:T}^v)=\mathcal{P}(x_T^v)\prod_{t=1}^T\mathcal{P}_\theta(x_{t-1}^v|x_t^v).
\end{equation} 
 
DiffCLIP acquires the clean input data $x_0$ through two parallel unsupervised learning constraints. Initially, it iteratively enhances the predicted scores of the unmasked $x_0^v$ during denoising restoration. Additionally, it restores the masked $x_0^m$ by formulating a set of learnable parameters equivalent to the number of masked patches. In the end, executing through a modality-shared image encoder and two modality-specific decoders, we update the DiffCLIP $f(\cdot)$ by minimizing the denoising restoration loss:
\begin{equation} \label{3.6}
    \mathbb{E}_{x_0\sim \mathcal{Q}(x_0)}\mathbb{E}_{x_T\sim \mathcal{N}(0,I)}||x_0-f(x_T, t)||^2.
\end{equation} 

\subsubsection{Encoder and Decoders}
The encoder uses standard ViT \cite{dosovitskiy2020image} and embeds trainable parameters on the diffused visible patches before encoding:
\begin{equation} \label{3.7}
    \hat{x}^v=\mathcal{E}_\theta\left(\{p_{ct}+p_p+p_{dt};x^v\}_1\right),
\end{equation} 
where $\{;\}_1$ means performing concatenate operation in the first dimension. $\mathcal{E}_\theta(\cdot)$ denotes modality-shared image encoder, $p_{ct}$, $p_p$, and $p_{dt}$ represent class tokens, positional embedding and diffusion timestep, respectively. It is noteworthy that each modality undergoes its linear projection layer after being mapped to the shared subspace to match dimensions.

To restore the real data of different modalities, we designed two decoders that, compared to the encoder, are more lightweight but fully capable of performing this task. Each decoder takes in visible tokens $\hat{x}^v$ and masked tokens $\hat{x}^m$ as the input. Each of the masked tokens is a learnable vector initialized to zeros. Similar to the encoder, we embed trainable positional embedding into the input data before decoding:
\begin{equation} \label{3.8}
    \hat{x}=\mathcal{D}_\theta\left(\{p_p^v+\hat{x}^v;p_p^m+\hat{x}^m\}_1\right),
\end{equation} 
where $\hat{x}$ represents the reconstructed data restored to the original space and $\mathcal{D}_\theta(\cdot)$ means modality-specific decoder. The introduced diffusion model creates a challenging yet effective task to assist the decoder in self-restoration.

The unsupervised training process entails defining a hybrid optimization objective $\mathcal{L}_{\text {UMD }}$, which incorporates the denoising restoration loss of visible patches and the restoration loss of masked patches:
\begin{equation} \label{3.18}
    \mathcal{L}_{\text {UMD}}=\mathbb{E}_{x_{0}}\|x_{0}-\hat{x}\|^{2}.
\end{equation} 


\subsection{Few-shot Language-Driven Classification}
CLIP \cite{radford2021learning} is a widely used vision-language model that learns a joint embedding space between images $x$ and texts $y$ by training on $N$ paired image-text data $\left\{x_i, y_i\right\}_{i=1}^N$ using contrastive learning. In this setup, positive pairs (matching images and texts) are aligned, while negative pairs (non-matching) are separated, enabling CLIP to perform zero-shot predictions based on text prompts. However, in the context of high-dimensional multimodal remote sensing, the lack of sufficient image-text pairs makes it difficult to train a robust model for accurate classification. DiffCLIP addresses this challenge by employing an unsupervised mask diffusion pre-training process, which enables the training of an effective image encoder using unlabeled samples. The training objective of DiffCLIP involves two classification tasks \cite{mo2023s}: predicting a text given an image \(p(y \mid x)\) and predicting an image given a text \(p(x \mid y)\). Each sample in the batch is assigned a label corresponding to its paired data as the target.

\subsubsection{Text Description Generation} \label{sec3.2.1}
In this section, we set specific class text descriptions beyond simply appending semantically informative adjectives, and refrain from loading any pre-trained weights for the text encoder, to underscore the effectiveness of our method. Specifically, texts encompass inherent attributes, inter-class relationships, and class names by incorporating prior knowledge. Taking the class ``Healthy Grass" as an example, we prompt that the grass is predominantly green in color, possesses a fine texture, and exhibits uniform distribution. We also suggest that healthy grass typically thrives under large trees or alongside roads, rooted in soil. These expanded text descriptions for specific classes can establish strong semantic associations between text and visual features to ensure a consistent representation.
Before encoding, we obtain the tokenized representation of the text information through simple embedding. Subsequently, we utilize the transformer to encode the tokenized representation and produce text feature embedding, and the embedding is normalized to have a unit norm denoted as:
\begin{equation} \label{3.9}
    v_{\mathrm{text}}=\mathcal{T}_\theta(y),
    z_{\mathrm{text}}=v_{\mathrm{text}}/\|v_{\mathrm{text}}\|,
\end{equation} 
where $y$ corresponds to tokenized text inputs and $\mathcal{T}_\theta(\cdot)$ denotes the transformer, these text feature vectors serve as semantic information for text and are utilized to align with the image feature space.

\subsubsection{Visual Features Representation}
Our primary focus is on obtaining visual features. As previously mentioned, the modality-shared encoder, trained using unsupervised mask diffusion, serves as a replacement for the image encoder in CLIP. The parameters of this modality-shared encoder are loaded, with only the linear projection layer cascaded for dimensional reduction being randomly initialized, then they are fine-tuned together. We posit that the modality-shared encoder has effectively captured common information between the two modalities.
Moreover, it has been imbued with robust semantic cues through the denoising restoration task during the decoding process, while preserving each modality's distinctive features. The process is formulated as follows:
\begin{equation} \label{3.11}
    v_{\mathrm{image}}=\mathcal{E}_\theta(x),
    z_{\mathrm{image}}=v_{\mathrm{image}}/\|v_{\mathrm{image}}\|,
\end{equation} 
where $x$ corresponds to patched image inputs. $v_{\mathrm{image}}$ is the feature vector obtained by encoding the patched image, and it is normalized to get $z_{\mathrm{image}}$.

DiffCLIP utilizes a softmax function with a temperature parameter $\tau>0$ applied to the cosine similarity of embeddings to predict the label:
\begin{equation} \label{3.13}
p(x \mid y)=\sigma_\tau\left(z_{\text{text}},\left\{z_{\text{image}}^i\right\}_{i=1}^N\right) \in \mathbb{R}^N,
\end{equation} 
\begin{equation} \label{3.14}
p(y \mid x)=\sigma_\tau\left(z_{\text{image}},\left\{z_{\text{text}}^i\right\}_{i=1}^N\right) \in \mathbb{R}^N,
\end{equation} 
where the softmax classifier $\sigma_\tau$ represents a function of input and output embeddings, and probabilities are computed as follows: 
\begin{equation} \label{3.15}
    p\left(x=x_i \mid y\right)= \frac{\exp \left(z_{\text{text}} \cdot z_{\text{image}}^i / \tau\right) }{\sum_{j=1}^N \exp \left(z_{\text{text}} \cdot z_{\text{image}}^j / \tau\right)},
\end{equation} 

\begin{equation} \label{3.16}
    p\left(y=y_i \mid x\right)= \frac{\exp \left(z_{\text{image}} \cdot z_{\text{text}}^i / \tau\right)}{\sum_{j=1}^N \exp \left(z_{\text{image}} \cdot z_{\text{text}}^j / \tau\right)}.
\end{equation}
The DiffCLIP model minimizes the following objectives:
\begin{equation} \label{3.17}
\mathcal{L}_{\text {FLC}}=\frac{1}{2 N} \sum_{i=1}^N\left(H\left(p\left(y \mid x_i\right), \mathbf{e}_i\right)+H\left(p\left(x \mid y_i\right), \mathbf{e}_i\right)\right),
\end{equation} 
where $H$ denotes the cross-entropy loss and $\mathbf{e}_i \in \mathbb{R}^N$ is a one-hot vector with the $i$-th element being one.

\section{Experiments}
\subsection{Experiments Setup}
\subsubsection{Datasets Description}
The experiments are conducted on four widely recognized benchmarks to assess the performance of our proposed method: Houston \cite{debes2014hyperspectral}, Trento \cite{rasti2017hyperspectral}, MUUFL \cite{gader2013muufl} and MRNet dataset \cite{bien2018deep}.
\subsubsection{Evaluation Metric and Comparison Methods}
To evaluate the classification outcomes quantitatively, we employ three key metrics: overall accuracy (OA), average accuracy (AA), and Kappa coefficient. To assess the effectiveness of our method, we select four SOTA methods in multimodal remote sensing classification: GLT \cite{ding2022global}, CALC \cite{lu2023coupled}, MIViT \cite{zhang2024multimodal}, LDS$^2$AE \cite{qu2024lds2ae}, five SOTA few-shot learning algorithms: RN-FSC \cite{gao2020deep}, MFRN-ML \cite{dai2024learning}, SCFormer \cite{li2024scformer}, HIPL \cite{yin2024hierarchy}, MMPR \cite{wu2024fine}, and CLIP \cite{radford2021learning}.

\begin{figure}[htb]
    \centering
    \includegraphics[width=1\linewidth]{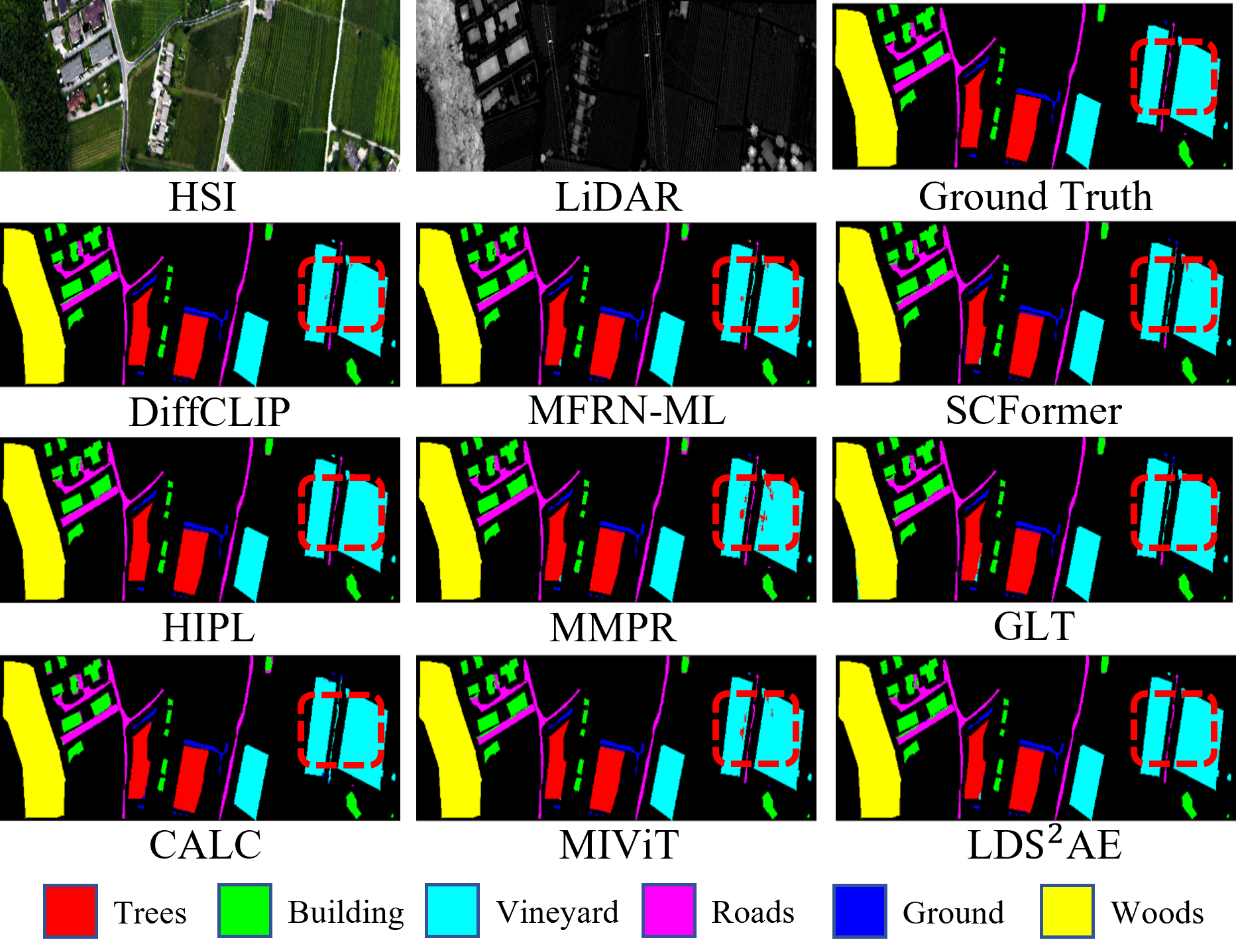}
    \caption{Classification maps of the Trento dataset.}
    \label{fig:Trento}
\end{figure}

\begin{table*}[htpb]
	\centering
 \vspace{-1pt}
    \setlength{\tabcolsep}{2mm}
    \begin{tabular}{c|c|ccc|ccc|ccc}
       \hline \hline
      \multirow{2}{*}{Settings}  &\multirow{2}{*}{Methods} & \multicolumn{3}{c|}{Houston} & \multicolumn{3}{c|}{MUUFL} & \multicolumn{3}{c}{Trento} \\ & & OA(\%) 
          & AA(\%)      & Kappa(\%)   & OA(\%)     & AA(\%)    & Kappa(\%)  & OA(\%)      & AA(\%)      & Kappa(\%)  \\ \hline \hline
  \multirow{7}{*}{2-shot}      &CLIP & 35.83 & 42.75 & 32.44 & 51.62 & 38.26 & 38.91 & 85.33 & 78.63 & 82.47  \\
        &LDS$^2$AE  & 48.68 & 51.68 & 44.85 & 56.47 & 44.86 & 43.78 & 90.23 & 79.25 & 86.76  \\
        & MFRN-ML  & 48.73 & 51.55 & 43.79 & 57.16 & 44.54 & 43.12 & 90.81 & 79.33 & 86.04  \\
        &SCFormer  & 49.09 & 52.93 & 46.11 & 57.37 & 46.94 & 45.13 & 91.55 & 80.05 & 88.24  \\ 
        &HIPL   & 50.96 & 54.33 & 47.15 & 58.84 & 52.16 & 47.37 & 92.11 & 80.14 & 88.59  \\
        &MMPR  & 49.79 & 53.01 & 46.78 & 57.86 & 49.05 & 46.15 & 91.75 & 79.88 & 87.93  \\
        &DiffCLIP  & \textbf{52.15} & \textbf{56.02} & \textbf{48.39} & \textbf{59.39} & \textbf{54.94} & \textbf{48.80} & \textbf{93.19} & \textbf{80.45} & \textbf{90.85}  \\ 
        \hline
   \multirow{7}{*}{8-shot}     &CLIP  & 52.11 & 56.37 & 52.86 & 67.56 & 66.78 & 63.21 & 92.76 & 91.66 & 92.53 \\ 
        &LDS$^2$AE  & 57.72 & 57.60 & 54.36 & 70.59 & 68.43 & 62.94 & 95.49 & 95.62 & 94.03  \\ 
        &MFRN-ML  & 58.03 & 60.59 & 55.71 & 72.36 & 70.28 & 65.39 & 95.54 & 95.78 & 94.02  \\  
        &SCFormer  & 59.61 & 60.92 & 56.25 & 74.38 & 70.64 & 67.82 & 95.57 & 95.79 & 94.18  \\
        &HIPL   & 60.36 & 63.89 & 57.54 & 75.65 & 72.63 & 69.15 & 95.98 & 95.91 & 94.79  \\
        &MMPR  & 59.98 & 62.57 & 56.92 & 74.67 & 71.68 & 67.98 & 95.63 & 95.72 & 94.33  \\
        &DiffCLIP  & \textbf{61.93} & \textbf{65.93} & \textbf{58.95} & \textbf{77.60} & \textbf{74.97} & \textbf{71.53} & \textbf{96.30} & \textbf{96.22} & \textbf{95.09}  \\    
        \hline
   \multirow{7}{*}{20-shot}     &CLIP  & 56.31 & 59.98 & 54.32 & 67.30 & 71.79 & 65.10 & 91.28 & 93.11 & 92.43  \\
        &LDS$^2$AE   & 65.71 & 70.50 & 63.09 & 78.13 & 77.24 & 72.01 & 98.13 & 96.91 & 97.50  \\ 
        &MFRN-ML   & 74.49 & 77.81 & 73.22 & 78.83 & 77.45 & 72.32 & 98.25 & 96.98 & 97.43  \\ 
        &SCFormer   & 74.97 & 78.16 & 73.80 & 79.11 & 78.55 & 73.23 & 98.27 & 96.95 & 97.36  \\
        &HIPL   & 79.36 & 82.11 & 76.35 & 80.87 & 79.36 & 76.28 & 98.41 & 97.15 & 97.86  \\
        &MMPR  & 77.93 & 80.26 & 75.70 & 80.02 & 78.71 & 74.63 & 98.40 & 97.03 & 97.42  \\
        &DiffCLIP   & \textbf{81.87} & \textbf{84.09} & \textbf{80.40} & \textbf{81.81} & \textbf{80.67}& \textbf{76.64} & \textbf{98.60} & \textbf{97.90} & \textbf{98.13} \\ 
        \hline 
        \hline
\end{tabular}

	\caption{Comparison results on the three datasets under various few-shot settings.}
\vspace{-0.1in}
\label{tab:4.3}
\end{table*}
\subsubsection{Implementation Details}
The experiments are conducted on a system with an NVIDIA GeForce RTX A100 GPU. During preprocessing, training samples are cropped into $11 \times 11$ patches. For optimization in both unsupervised and few-shot learning, the Adam optimizer is used with an initial learning rate of 1e-4 and weight decay of 1e-5. Two schedulers are employed: a cosine scheduler for unsupervised learning and a step scheduler for few-shot learning. The training consists of 100 epochs for unsupervised learning and 150 epochs for few-shot learning. To ensure optimal performance in comparative experiments, the batch size is set to 256 for unsupervised learning and 64 for few-shot learning, with consistent parameter settings across all datasets.

\subsection{Comparison Results}
To evaluate our proposed method and compare it with current SOTA methods in the few-shot learning classification task, we conduct experiments on diverse datasets: Houston, Trento, and MUUFL. As shown in Table \ref{tab:4.3}, 
DiffCLIP consistently outperforms other models in three datasets, showing improved performance as the number of training images increases. By leveraging unsupervised mask diffusion, our method effectively learns modality-specific and shared features across modalities while capturing individual modality data distributions. Moreover, through semantic supervision and modality-specific feature enhancement via denoising restoration tasks, our method effectively extracts multimodal information even with limited training samples and achieves an overall accuracy improvement of 10.65\% across three datasets compared with CLIP, highlighting DiffCLIP’s few-shot learning capabilities.
We also conduct qualitative evaluations by visualizing the classification maps on the Trento dataset in Figure \ref{fig:Trento}. DiffCLIP achieves optimal visual classification performance, effectively utilizing semantic information, particularly demonstrating perfect classification continuity in categories such as Roads.

\begin{table}[htpb]
	\centering
	 \scalebox{1}{
\begin{tabular}{c|ccc}
\hline \hline
Sample numbers    & OA(\%)    & AA(\%)    & Kappa(\%)  \\ \hline \hline
300      & 91.75 & 91.11 & 91.62  \\
500 & 92.58 & 91.76 & 93.00  \\
700    &\textbf{94.80} &\textbf{94.24} &\textbf{93.28}   \\ \hline \hline
\end{tabular}}
	\caption{Ablation of sample numbers used in unsupervised mask diffusion stage on three datasets.}

\vspace{-0.2in}
\label{tab:samples}
\end{table}

\subsection{Ablation Studies}
\subsubsection{Ablation of Sample Numbers}
The impact of sample size in the unsupervised mask diffusion stage is evaluated by varying the number of samples from 300 to 700 in Table \ref{tab:samples}. Results show that increasing the sample size leads to notable improvements in OA, AA, and Kappa coefficient. Specifically, the model achieves its best performance with 700 samples. This indicates that a larger sample size enhances the model’s classification accuracy and reliability, with 700 samples offering an optimal balance between performance and efficiency.

\subsubsection{Ablation of Parameter Settings}
\begin{figure}[htpb]
    \centering
    \includegraphics[width=1\linewidth]{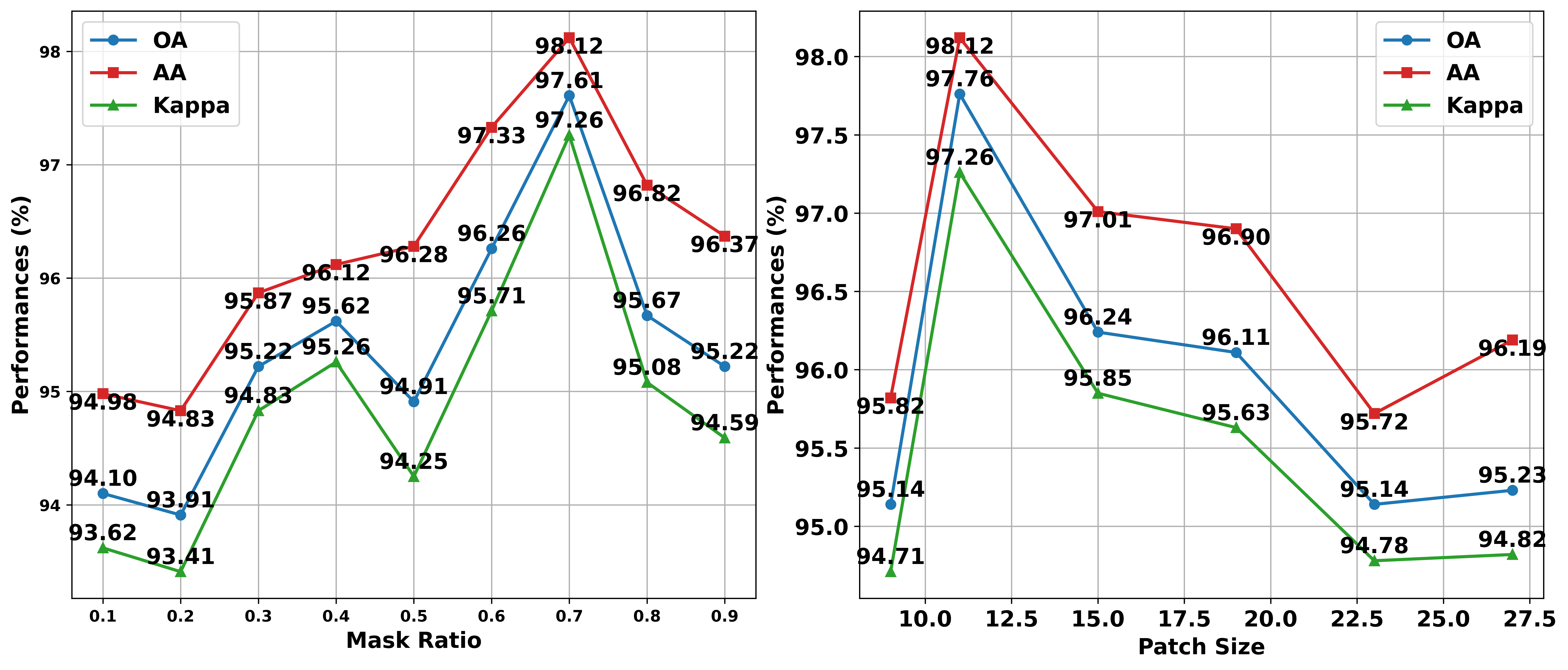}
    \caption{Classification performance of Houston dataset with different masking ratios and patch size.}
    \label{fig:para}
    \vspace{-0.2in}
\end{figure}

The study in Figure \ref{fig:para} on DiffCLIP’s performance explores the effects of masking ratios and patch sizes. Increasing the masking ratio initially improves performance by reducing redundancy, but excessive masking leads to information loss. Results show a peak at 70\%, which is selected for further experiments. Similarly, ablation experiments reveal that a patch size of 11 performs best, balancing category separation and receptive field size for optimal results.

\subsubsection{Ablation of Model Architecture}
The study in Table \ref{tab:4.2} evaluates DiffCLIP by comparing it to four baseline approaches: i) replacing the text encoder with a fully connected layer, ii) removing the masking process, iii) removing the diffusion process, and iv) eliminating the entire unsupervised mask diffusion process. Results show that removing the text encoder reduces OA by 5.45\%, underscoring the importance of textual information for semantics. Removing the diffusion and masking processes decreases OA by 5.13\% and 4.02\%, respectively, highlighting their role in model robustness. Eliminating the entire unsupervised mask diffusion leads to a significant 6.98\% OA drop, affirming its critical role in adapting CLIP to remote sensing. Overall, DiffCLIP consistently outperforms the baselines.

\begin{table}[htpb]
	\centering
	 \scalebox{1}{
\begin{tabular}{c|ccc}
\hline \hline
 Settings             & OA(\%)    & AA(\%)    & Kappa(\%)  \\ \hline \hline
w/o Text      & 89.35 & 89.66 & 89.01  \\
w/o Diffusion & 89.67 & 90.73 & 89.28  \\
w/o Mask & 90.78 & 91.72 & 89.83  \\
w/o Unsupervised & 87.82 & 88.94 & 87.76  \\
DiffCLIP    &\textbf{94.80} &\textbf{94.24} &\textbf{93.28}   \\ \hline \hline
\end{tabular}}
\caption{Ablation of model components on three datasets.}
\vspace{-0.2in}
\label{tab:4.2}
\end{table}

\begin{table*}[ht]
\centering
  \vspace{-0.1in}
    \setlength{\tabcolsep}{0.7mm}
\begin{tabular}{c|c|c|ccc}
\hline \hline
  & Context length  & Text prompts                         & OA(\%)    & AA(\%)    & Kappa(\%) \\
   \hline \hline
p1 & 5$\sim$7            & a patch of a \{class name\}.                   & 97.61 & 98.12 & 97.26 \\
p2 & 6$\sim$8            & a nice patch of a \{class name\}.      & 97.05 & 97.63 & 96.81 \\
p3 & 7$\sim$9            & a fusion patch of a \{class name\}. & 97.63 & 98.11 & 97.34 \\
p4 & 11$\sim$13           & a multimodal fusion patch of a \{class name\} with strong semantic information.  & 97.93 & 98.51 & 97.64 \\
p5 & 15$\sim$30          & specific class text descriptions.               & \textbf{98.15} & \textbf{98.87} & \textbf{98.09} \\
\hline \hline
\end{tabular}
\caption{Ablation experiments of different text prompts and context length of Houston2013 dataset.}
\label{tab:4.4}
\end{table*}

\begin{table*}[tpb]
	\centering
    \setlength{\tabcolsep}{1.8mm}
\begin{tabular}{c|c|ccc|ccc|ccc}
\hline \hline
\multicolumn{2}{c|}{\multirow{2}{*}{Methods}} & \multicolumn{3}{c|}{Houston} & \multicolumn{3}{c|}{MUUFL} & \multicolumn{3}{c}{Trento} \\
\multicolumn{1}{c}{}            &             & OA(\%)      & AA(\%)      & Kappa(\%)   & OA(\%)     & AA(\%)    & Kappa(\%)  & OA(\%)      & AA(\%)      & Kappa(\%)  \\ \hline \hline
\multirow{4}{*}{Supervised} 
& GLT     & 90.13   & 90.42   & 89.42   & 82.75   & 75.70   & 78.67  & 98.19   & 97.75   & 98.04  \\
& CALC               & 87.87   & 88.92   & 86.87   & 81.94   & 64.09  & 77.01  & 97.11   & 92.31   & 96.64  \\
& MIViT        & 93.21   & 93.87   & 92.75   & 83.13   & 79.74  & 78.91  & 98.03   & 97.96   & 98.24  \\
& LDS$^2$AE       & 94.88   & 95.31   & 94.46   & 84.82   & 82.19  & 80.42  & 98.77   & 98.11   & 98.42  \\ \hline
\multirow{6}{*}{Few-shot} 
& RN-FSC    & 93.42   & 94.53   & 93.64   & 84.83	   & 74.52  & 81.25  & 98.69   & 97.66   & 98.61	\\ 
&MFRN-ML   & 94.50 & 95.38 & 93.99 & 84.78 & 77.03 & 81.07 & 98.37 & 97.71 & 98.49  \\ 
&SCFormer   & 95.96 & 96.25 & 94.84 & 84.93 & 79.49 & 81.39 & 98.52 & 97.86 & 98.55  \\
&HIPL   & 96.83 & 97.01 & 96.75 & 85.45 & 80.71 & 82.18 & 98.81 & 98.15 & 98.65  \\
&MMPR  & 96.54 & 96.77 & 95.98 & 85.07 & 80.36 & 81.57 & 98.43 & 97.79 & 98.46  \\
& DiffCLIP     &\textbf{98.15} &\textbf{98.87} &\textbf{98.09}   & \textbf{86.98}   & \textbf{85.01}  & \textbf{82.81}  & \textbf{99.26}   & \textbf{98.84}   & \textbf{98.95}  \\ \hline \hline
\end{tabular}
\caption{Comparison results of OA, AA, and Kappa on three datasets.}
\vspace{-10pt}
\label{tab:4.1}
\end{table*}

\subsubsection{Ablation of Text Prompts}
DiffCLIP’s classification performance benefits from language-driven methods. To assess the impact of text prompts, five crafted prompt sets are tested on the Houston dataset. The baseline, p1, is compared against extended prompts. As seen in Table \ref{tab:4.4}, p2’s longer text reduces accuracy due to weaker task relevance, while p3’s minor adjustments with effective “fusion” text achieve minimal improvement. p4 demonstrates that longer, relevant prompts improve supervision. p5’s specific class descriptions boosted OA accuracy by 0.54\%. This highlights the importance of optimizing both prompt length and content for better performance in DiffCLIP.

\begin{figure}[htpb]
    \centering
    \includegraphics[width=\linewidth]{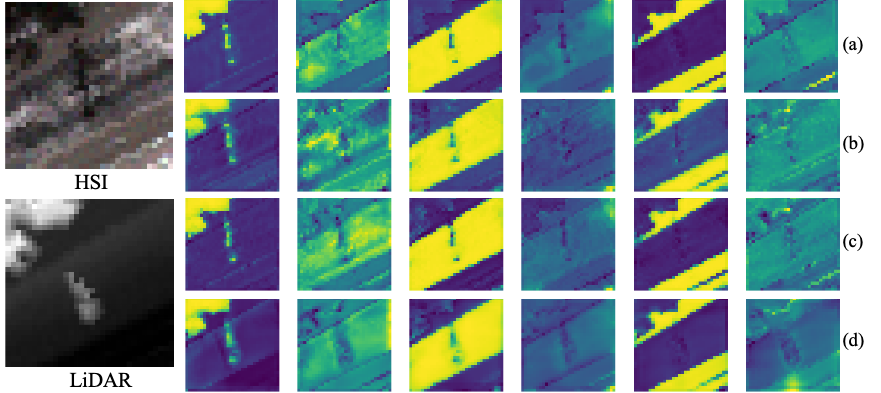}
    \caption{Feature visualization of the visual encoder.}
    \vspace{-6mm}
    \label{features}
\end{figure}

\subsection{Feature Visualization}
The few-shot learning is based on the pre-trained visual encoder obtained from the unsupervised mask diffusion. On this basis, the text encoder is added, and a supervised learning process is conducted with a mere number of paired images and texts (\textit{e.g.} 2 shot), which utilizes a limited number of labeled samples to train the model. We plot the visual encoder feature visualization of different modalities in Figure \ref{features}. (a) and (c) are from unsupervised mask diffusion, (b) and (d) are from few-shot language-driven classification. This similar visualization highlights the excellent performance of DiffCLIP for domain-invariant feature extraction and eliminates the domain shift. 

\begin{table}[htpb]	
    \centering
    \scalebox{1}{
    \begin{tabular}{c|cccc}
        \hline \hline
        Method  &ACC &AUC &SE &SP \\
        \hline \hline
        ELNet & 0.639 & 0.703&0.624&0.650\\
        TransMed-S &0.667&0.705&0.635&0.664\\
        SSL-DcGaR &0.731&0.758&0.723&0.734\\
        \textbf{DiffCLIP}&\textbf{0.763}&\textbf{0.787}&\textbf{0.750}&\textbf{0.756}\\
      \hline \hline
    \end{tabular}}
    \caption{Comparison results of the MRNet dataset.}
    \vspace{-0.2in}
     \label{tab3}
\end{table}

\subsection{Generalization Validation}
We also compare with fully supervised algorithms in Tabel \ref{tab:4.1}. For fair comparison, we randomly sample 40 samples per class for training with labels, and the remaining samples for evaluation. Our approach consistently achieves the highest scores across all datasets and evaluation metrics, including OA, AA, and Kappa. To validate the generalization of our method to the other fields, we compare the DiffCLIP with ELNet \cite{wu2021elnet}, TransMed-S \cite{dai2021transmed}, and SSL-DcGaR \cite{berenguer2024semi}, using 10 samples of MRNet data to train and the rest to test. As shown in Table \ref{tab3}, even in the multimodal medical field, DiffCLIP still maintains better performance.

\section{Conclusion}
In summary, our proposed DiffCLIP introduces a novel paradigm for high-dimensional multimodal few-shot remote sensing image classification, addressing the challenge posed by limited training samples in this domain. DiffCLIP utilizes unsupervised mask diffusion pre-training to transform multimodal data into a shared subspace for learning multimodal shared features. It maintains modality-specific features through modality-specific decoders and operates effectively with a small number of training samples. This method provides a robust image feature encoder for few-shot learning while also reducing computational costs. An essential aspect of DiffCLIP is its utilization of language-driven methods to convey comprehensive semantic information. This aids the model in capturing rich inherent semantic details in complex data distributions. By integrating text features with visual features, DiffCLIP facilitates the fusion of multimodal information, leading to improved classification performance. Our results on benchmark datasets demonstrate the effectiveness and robustness of the proposed method.

\bibliography{aaai25}

\end{document}